\documentclass[11pt,a4paper]{article}
\usepackage[hyperref]{acl2020}
\usepackage{threeparttable}
\usepackage{subcaption}
\usepackage{graphicx}
\usepackage{times}
\usepackage{latexsym}
\usepackage{tablefootnote}
\usepackage{url}

\usepackage{booktabs}
\usepackage{adjustbox}

\usepackage{amsmath}
\usepackage{amsfonts}
\usepackage{bbm}
\usepackage{bm}
\usepackage{mathtools}
\usepackage{microtype}
\usepackage{scalerel} %

\usepackage{tikz}
\usetikzlibrary{shapes, positioning, fit, calc}
\usetikzlibrary{backgrounds}
\usetikzlibrary{arrows}
\usetikzlibrary{arrows.meta}
\usetikzlibrary{shapes.arrows, shapes.geometric}
\usetikzlibrary{matrix}

\usepackage{pgfplots}
\usepgfplotslibrary{groupplots}
\pgfplotsset{compat=1.14}

\usepackage{xcolor}

\definecolor{turmeric}{RGB}{204,203,77} 

\definecolor{oyster_pink}{RGB}{238,206,205} 
\definecolor{coral_candy}{RGB}{242,208,205} 
\definecolor{baby_pink}{RGB}{246, 194, 192}
\definecolor{oyster_pink}{RGB}{238,206,205} 
\definecolor{NY_pink}{RGB}{228,136,113} 
\definecolor{petite_orchid}{RGB}{223, 157, 155}
\definecolor{carmine_pink}{RGB}{231, 76, 60}
\definecolor{deep_carmine_pink}{RGB}{236, 50, 67}
\definecolor{apricot}{RGB}{241,140,122}

\definecolor{milan}{RGB}{255, 254, 169}
\definecolor{casablanca}{RGB}{244, 178, 84}
\definecolor{texas}{RGB}{245, 232, 123}
\definecolor{maize}{RGB}{249, 212, 156}

\definecolor{double_pearl_lusta}{RGB}{253, 242, 208}

\definecolor{oasis}{RGB}{253, 242, 208}
\definecolor{linen}{RGB}{251, 239, 227}

\definecolor{zanah}{RGB}{220, 233, 213}
\definecolor{frostee}{RGB}{217, 231, 214}
\definecolor{norway}{RGB}{158, 194, 132}
\definecolor{malibu}{RGB}{110, 180, 240}

\definecolor{link_water}{RGB}{221, 232, 250}

\definecolor{spring_leaves}{RGB}{46, 83, 117}

\definecolor{venice_blue}{RGB}{87, 135, 105}
\definecolor{boston_blue}{RGB}{68, 147, 161}

\definecolor{napa}{RGB}{163, 154, 137}

\definecolor{mexican_red}{RGB}{170, 41, 37}
\definecolor{valencia}{RGB}{214, 87, 70}

\definecolor{riptide}{RGB}{141,211,199}
\definecolor{pale_prim}{RGB}{255,255,179}
\definecolor{lavender_gray}{RGB}{190,186,218}
\definecolor{salmon}{RGB}{242,131,107}
\definecolor{seagull}{RGB}{128,177,211}
\definecolor{rajah}{RGB}{253,180,98}
\definecolor{yellow_green}{RGB}{198,222,119}
\definecolor{classic_rose}{RGB}{252,205,229}
\definecolor{feijoa}{RGB}{178,223,138}

\definecolor{cruise}{RGB}{179,226,205}

\definecolor{periwinkle}{RGB}{203,213,232}
\definecolor{snow_flurry}{RGB}{230,245,201}
\definecolor{buttermilk}{RGB}{255,242,174}

\definecolor{sundown}{RGB}{249, 180, 181}
\definecolor{spindle}{RGB}{179,205,227}
\definecolor{tea_green}{RGB}{204,235,197}
\definecolor{languid_lavender}{RGB}{222,203,228}
\definecolor{champagne}{RGB}{254,217,166}
\definecolor{cream}{RGB}{255,255,204}

\definecolor{monte_carlo}{RGB}{135,204,194}
\definecolor{melon}{RGB}{254,191,181}
\definecolor{granny_smith_apple}{RGB}{150,214,150}
\definecolor{watusi}{RGB}{254,221,207}
\definecolor{see_green}{RGB}{161,228,195}

\definecolor{moss_green}{RGB}{170,216,176}
\definecolor{opal}{RGB}{164,207,190}

\definecolor{pale_turquoise}{RGB}{172,240,242}
\definecolor{Madang}{RGB}{190,235,159}
\definecolor{pixie_green}{RGB}{183,214,170}
\definecolor{coral_andy}{RGB}{243,204,205}
\definecolor{manhattan}{RGB}{226,180,125}
\definecolor{quartz}{RGB}{219,223,238}
\definecolor{spring_sun}{RGB}{242,243,195}
\definecolor{dairy_cream}{RGB}{254,226,189}
\definecolor{surf_crest}{RGB}{205,230,208}
\definecolor{french_pass}{RGB}{195,232,246}
\definecolor{cosmos}{RGB}{248,209,210}
\definecolor{portafino}{RGB}{245,237,160}
\definecolor{sail}{RGB}{163,205,235}
\definecolor{hint_green}{RGB}{226,246,209}

\definecolor{jet_stream}{RGB}{188, 214, 210}

\definecolor{azalea}{RGB}{251, 196, 196}
\definecolor{wewak}{RGB}{244, 143, 150}
\definecolor{bittersweet}{RGB}{255,111,105}
\definecolor{sunset_orange}{RGB}{242,89,75}
\definecolor{light_coral}{RGB}{244, 127, 123}
\definecolor{carnation}{RGB}{245, 80, 86}
\definecolor{flamingo}{RGB}{237, 88, 85}

\definecolor{fire_engine_red}{RGB}{210,44,41}
\definecolor{amaranth}{RGB}{234,46,73}
\definecolor{ku_crimson}{RGB}{243, 0, 25}
\definecolor{fire_engine_red}{RGB}{206, 37, 51}
\definecolor{copper_rust}{RGB}{155, 64, 74}

\definecolor{chilean_fire}{RGB}{215, 87, 44}

\definecolor{japanese_laurel}{RGB}{53, 116, 40}

\definecolor{turmeric}{RGB}{211, 178, 76}
\definecolor{saffron}{RGB}{249,193,62}
\definecolor{my_sin}{RGB}{255, 176, 59}
\definecolor{tree_poppy}{RGB}{246, 154, 27}
\definecolor{jaffa}{RGB}{240, 131, 58}
\definecolor{crusta}{RGB}{254, 127, 44}
\definecolor{tahiti_gold}{RGB}{223, 102, 36}
\definecolor{outrageous_orange}{RGB}{255, 100, 45}
\definecolor{safety_orange}{RGB}{254, 106, 0}

\definecolor{turquoise}{RGB}{41,217,194}
\definecolor{puerto_rico}{RGB}{94, 194, 166}
\definecolor{mountain_meadow}{RGB}{0, 163, 136}
\definecolor{free_speech_aquamarine}{RGB}{0, 156, 114}
\definecolor{java}{RGB}{2,190,196}

\definecolor{matisse}{RGB}{25, 104, 167}
\definecolor{shakespeare}{RGB}{85, 154, 193}
\definecolor{mona_lisa}{RGB}{246,152,134}

\definecolor{bgc}{RGB}{245,245,245}
\definecolor{tuatara}{RGB}{67, 67, 67}
\definecolor{aluminum}{RGB}{153,153,153}
\definecolor{silver}{RGB}{191,191,191}
\definecolor{platinum}{RGB}{228,228,228}
\definecolor{mercury}{RGB}{230,230,230}
\definecolor{gallery}{RGB}{240,240,240}
\definecolor{athens_gray}{RGB}{236, 240, 241}
\definecolor{ship_gray}{RGB}{77,77,77}

\definecolor{early_dawn}{RGB}{252,243,218}
\definecolor{egg_shell}{RGB}{238, 234, 215}
\definecolor{midnight}{RGB}{0, 29, 50}
\definecolor{sundown}{RGB}{249, 180, 181}
\definecolor{sun_shade}{RGB}{255, 144, 68}
\definecolor{sushi}{RGB}{117, 168, 47}
\definecolor{tomato}{RGB}{255, 97, 56}
\definecolor{ice_cold}{RGB}{169,232,220}

\definecolor{jelly_bean}{RGB}{45, 126, 150}
\definecolor{celestial_blue}{RGB}{52, 152, 219}
\definecolor{curious_blue}{RGB}{41, 128, 185}
\definecolor{french_blue}{RGB}{0, 112, 182}
\definecolor{matisse}{RGB}{25, 104, 167}

\definecolor{biscay}{RGB}{44, 62, 80}

\definecolor{cosmic_latte}{RGB}{222, 247, 229}
\definecolor{chinook}{RGB}{163, 232, 178}
\definecolor{padua}{RGB}{121, 189, 143}
\definecolor{ocean_green}{RGB}{79, 176, 112}
\definecolor{pastel_green}{RGB}{107, 227, 135}
\definecolor{chateau_green}{RGB}{69, 191, 85}
\definecolor{RoyalBlue}{RGB}{69, 191, 85}
\definecolor{pigment_green}{RGB}{0, 175, 79}
\definecolor{fern}{RGB}{101,197,117}
\definecolor{killarney}{RGB}{56, 113, 66}
\definecolor{viridian}{RGB}{70, 137, 102}

\aclfinalcopy %

\title{CLEAR: Contrastive Learning for Sentence Representation}

\author{Zhuofeng Wu$^{1}$\thanks{\hspace{1.5 mm} Work done while the author was an intern at Facebook AI.}\hspace{5 mm} 
Sinong Wang$^2
$\hspace{5 mm} Jiatao Gu$^2$ \\\textbf{Madian Khabsa$^{2}$\hspace{7.5 mm} Fei Sun$^{3}$\hspace{12 mm} Hao Ma$^{2}$}\\
$^1$School of Information, University of Michigan \\ \tt{zhuofeng@umich.edu} \\
$^2$Facebook AI\\ \tt{\{sinongwang, jgu, mkhabsa, haom\}@fb.com} \\
$^3$Institute of Computing Technology, Chinese Academy of Sciences \\
\tt{ofey.sunfei@gmail.com}}
\date{}

\begin{document}
\maketitle
\begin{abstract}
Pre-trained language models have proven their unique powers in capturing implicit language features. 
However, most pre-training approaches focus on the word-level training objective, while sentence-level objectives are rarely studied. 
In this paper, we propose Contrastive LEArning for sentence Representation (CLEAR), which employs multiple sentence-level augmentation strategies in order to learn a noise-invariant sentence representation. These augmentations include word and span deletion, reordering, and substitution. 
Furthermore, we investigate the key reasons that make contrastive learning effective through numerous experiments.
We observe that different sentence augmentations during pre-training lead to different performance improvements on various downstream tasks.%
Our approach is shown to outperform multiple existing methods on both SentEval and GLUE benchmarks.
\end{abstract}

\section{Introduction} \label{introduction}

\tikzset{
  FARROW/.style={arrows={-{Latex[length=1.25mm, width=1.mm]}}, thick},
  origin/.style = {regular polygon, regular polygon sides=6, fill=silver!38!platinum, minimum width=2em, align=center, inner sep=-0.5mm, outer sep=0, font=\tiny, rounded corners=1},
  aug1/.style = {regular polygon, regular polygon sides=6, fill=monte_carlo, minimum width=2em, align=center, inner sep=-0.5mm, outer sep=0, font=\tiny, rounded corners=1},
  aug2/.style = {regular polygon, regular polygon sides=6, fill=Madang!150, minimum width=2em, align=center, inner sep=-0.5mm, outer sep=0, font=\tiny, rounded corners=1},
  emb/.style = {rectangle, fill=flamingo!72, minimum width=1.8em, minimum height=1.8em, align=center, inner sep=0, outer sep=0, rounded corners=1, font=\scriptsize},
  encoder/.style = {rectangle, fill=Madang!82, minimum width=8em, minimum height=3em, align=center, rounded corners=3},
  project/.style = {rectangle, fill=casablanca!82, minimum width=1.8em, minimum height=1.8em, align=center, rounded corners=1, inner sep=0, outer sep=0, font=\scriptsize},
}

\begin{figure*}
\centering
\resizebox{0.9\linewidth}{!}{
    \begin{tikzpicture}

    \node[] (o) at (0, 0) {Original sentence $s$};
    \node [above of=o, node distance=0.8cm, align=center] (w_dots) {\scriptsize $\cdots$};

    \node[origin, left of=w_dots, node distance=0.7cm] (w_i){Tok$_i$};
    \node[left of=w_i, node distance=0.7cm, align=center] (wd1){\scriptsize $\cdots$};
    \node[origin, left of=wd1, node distance=0.7cm] (w_1){Tok$_1$};
     \node[origin, right of=w_dots, node distance=0.7cm] (w_j){Tok$_j$};
    \node[right of=w_j, node distance=0.7cm, align=center] (wdr){\scriptsize $\cdots$};
    \node[origin, right of=wdr, node distance=0.6cm] (w_n){Tok$_{\scaleto{N}{2.6pt}}$};
    
    \node [thick, dotted, draw=black, fit={(w_1) (w_n)}, inner sep=3] (input) {};

    \node [above of=w_dots, node distance=2cm, xshift=-4.5cm] (a1_dots) {\scriptsize $\cdots$};
    \node [aug1, left of=a1_dots, node distance=0.7cm] (ai) {$\mathrm{Tok}_i'$};
    \node [left of=ai, node distance=0.7cm] (ad1) {\scriptsize $\cdots$};
    \node [aug1, left of=ad1, node distance=0.7cm] (a1) {$\mathrm{Tok}_1'$};
    \node [aug1, left of=a1, node distance=1.2cm] (a_cls) {[CLS]};
    \node [aug1, right of=a1_dots, node distance=0.7cm] (aj) {$\mathrm{Tok}_j'$};
    \node [right of=aj, node distance=0.7cm] (adr) {\scriptsize $\cdots$};
    \node [aug1, right of=adr, node distance=0.7cm, inner sep=-0.6mm] (an) {$\mathrm{Tok}_{\scaleto{N}{2.4pt}}'$};
    
     \foreach \source in {a1_dots, ad1, adr}
      {
        \node [above of=\source, node distance=1.5cm] (ed\source) {\scriptsize $\cdots$};
        \node [above of=\source, node distance=3.5cm] (pd\source) {\scriptsize $\cdots$};
      }
    
    \node [emb, above of=a_cls, node distance=1.5cm] (ea_cls) {$\bm{E}_{\scaleto{\mathrm{[CLS]}}{3pt}}$};
    \node [emb, above of=a1, node distance=1.5cm] (ea1) {$\bm{E}_1$};
     \node [emb, above of=ai, node distance=1.5cm] (eai) {$\bm{E}_{i}$};
     \node [emb, above of=aj, node distance=1.5cm] (eaj) {$\bm{E}_j$};
     \node [emb, above of=an, node distance=1.5cm] (ean) {$\bm{E}_N$};
     
     \node [project, above of=ea1, node distance=2cm] (ha1) {$\bm{h}_1$};
     \node [project, above of=eai, node distance=2cm] (hai) {$\bm{h}_i$};
     \node [project, above of=eaj, node distance=2cm] (haj) {$\bm{h}_j$};
     \node [project, above of=ean, node distance=2cm] (han) {$\bm{h}_{\scaleto{N}{2.4pt}}$};
     \node [project, above of=ea_cls, node distance=2cm] (ha_cls) {$\bm{h}_{\scaleto{\mathrm{[CLS]}}{2.6pt}}$};
     
     \begin{scope}[on background layer]
      \node [draw=black, fill=french_pass!72, fill opacity=1, fit={(ea_cls) (han)}, inner sep=0.7em, rounded corners=3, label={center:transformer encoder $f(\cdot)$}] (enc_a) {};
     \end{scope}
    
    \node [above of=ha_cls, node distance=1.8cm] (out_a) {$\bm{z}_1$};
    \node [thick, dotted, draw=black, fit={(a1) (an)}, inner sep=3] (input_a) {};
    \draw [thick,draw=black!42, decorate,decoration={brace,amplitude=8pt,mirror}] ([xshift=-2.5mm, yshift=-2.2mm] a1.south west) -- ([xshift=2.5mm, yshift=-2.2mm] an.south east) node[midway,yshift=-0.5cm] (a_a) {$\widetilde{s}_1=\mathrm{AUG}(s, \mathit{seed}_1)$};

    \node [above of=w_dots, node distance=2cm, xshift=4.5cm] (b1_dots) {\scriptsize $\cdots$};
    \node [aug2, left of=b1_dots, node distance=0.7cm] (bi) {$\mathrm{Tok}_i''$};
    \node [left of=bi, node distance=0.7cm] (bd1) {\scriptsize $\cdots$};
    \node [aug2, left of=bd1, node distance=0.7cm] (b1) {$\mathrm{Tok}_1''$};
    \node [aug2, left of=b1, node distance=1.2cm] (b_cls) {[CLS]};
    \node [aug2, right of=b1_dots, node distance=0.7cm] (bj) {$\mathrm{Tok}_j''$};
    \node [right of=bj, node distance=0.7cm] (bdr) {\scriptsize $\cdots$};
    \node [aug2, right of=bdr, node distance=0.7cm, inner sep=-0.6mm] (bn) {$\mathrm{Tok}_{\scaleto{N}{2.4pt}}''$};
    
    \foreach \source in {b1_dots, bd1, bdr}
      {
        \node [above of=\source, node distance=1.5cm] (ed\source) {\scriptsize $\cdots$};
        \node [above of=\source, node distance=3.5cm] (pd\source) {\scriptsize $\cdots$};
      }
      
    \node [emb, above of=b_cls, node distance=1.5cm] (eb_cls) {$\bm{E}_{\scaleto{\mathrm{[CLS]}}{3pt}}$};
    \node [emb, above of=b1, node distance=1.5cm] (eb1) {$\bm{E}_1$};
     \node [emb, above of=bi, node distance=1.5cm] (ebi) {$\bm{E}_i$};
     \node [emb, above of=bj, node distance=1.5cm] (ebj) {$\bm{E}_j$};
     \node [emb, above of=bn, node distance=1.5cm] (ebn) {$\bm{E}_N$};
     
     \node [project, above of=eb1, node distance=2cm] (hb1) {$\bm{h}_1$};
     \node [project, above of=ebi, node distance=2cm] (hbi) {$\bm{h}_i$};
     \node [project, above of=ebj, node distance=2cm] (hbj) {$\bm{h}_j$};
     \node [project, above of=ebn, node distance=2cm] (hbn) {$\bm{h}_{\scaleto{N}{2.4pt}}$};
     \node [project, above of=eb_cls, node distance=2cm] (hb_cls) {$\bm{h}_{\scaleto{\mathrm{[CLS]}}{2.6pt}}$};
     
     \begin{scope}[on background layer]
      \node [draw=black, fill=french_pass!72, fill opacity=1, fit={(eb_cls) (hbn)}, inner sep=0.7em, rounded corners=3, label={center:transformer encoder $f(\cdot)$}] (enc_b) {};
     \end{scope}
     \node [above of=hb_cls, node distance=1.8cm] (out_b) {$\bm{z}_2$};
     
     \node [thick, dotted, draw=black, fit={(b1) (bn)}, inner sep=3] (input_b) {};
     \draw [thick,draw=black!42, decorate,decoration={brace,amplitude=8pt,mirror}] ([xshift=-2.5mm, yshift=-2.2mm] b1.south west) -- ([xshift=2.5mm, yshift=-2.2mm] bn.south east) node[midway,yshift=-0.5cm] (b_a) {$\widetilde{s}_2=\mathrm{AUG}(s, \mathit{seed}_2)$};
     
      \foreach \source/\target in {b_cls/eb_cls, b1/eb1, bi/ebi, bj/ebj, bn/ebn, a_cls/ea_cls, a1/ea1, ai/eai, aj/eaj, an/ean}
      {
        \draw[FARROW]  (\source) ->  (\target)  {} ;
      }
      \draw[FARROW]  ([yshift=2.8mm] ha_cls.north) ->  (out_a) node[pos=0.5, right] {\scriptsize $g(\cdot)$} ;
      \draw[FARROW]  ([yshift=2.8mm] hb_cls.north) ->  (out_b) node[pos=0.5, right] {\scriptsize $g(\cdot)$} ;

      \node [right of=out_a, node distance=4.5cm] (loss) {maximize agreement};
      \draw[FARROW]  (out_a) ->  (loss.west)  {} ;
      \draw[FARROW]  (out_b) ->  (loss.east)  {} ;
      
       \draw[FARROW, rounded corners,] (input) -| (a_a.south) node[pos=0.24, sloped, above] {\scriptsize $\mathrm{AUG} \sim \mathcal{A}$} ;
       \draw[FARROW, rounded corners,] (input) -| (b_a.south) node[pos=0.24, sloped, above, yshift=-0.03cm] {\scriptsize $\mathrm{AUG} \sim \mathcal{A}$} ;

    \end{tikzpicture} }
    \caption{The proposed contrastive learning framework CLEAR.}
    \label{fig:cl_model}
\end{figure*}
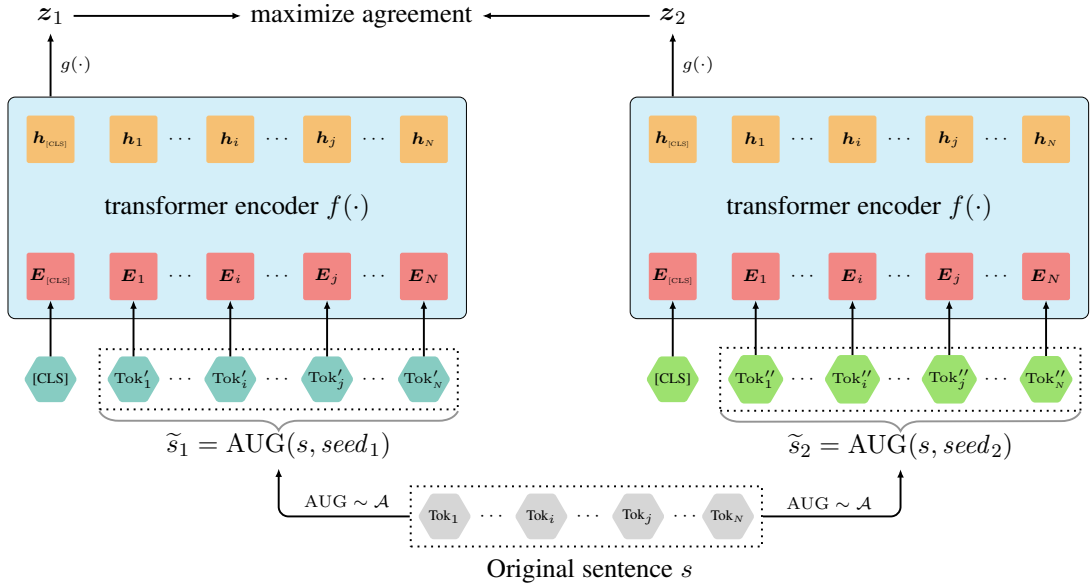

Learning a better sentence representation model has always been a fundamental problem in Natural Language Processing (NLP). Taking the mean of word embeddings as the representation of sentence (also known as mean pooling) is a common baseline in the early stage. Later on, pre-trained  models such as BERT~\cite{devlin2019bert} propose to insert a special token (i.e., [CLS] token) during the pre-training and take its embedding as the representation for the sentence. Because of the tremendous improvement brought by BERT~\cite{devlin2019bert}, people seemed to agree that CLS-token embedding is better than averaging word embeddings. Nevertheless, a recent paper Sentence-BERT~\cite{reimers2019sentence} observed that averaging of all output word vectors outperforms the CLS-token embedding marginally. Sentence-BERT's results suggest that models like BERT learn a better representation at the token level. One natural question is how to better learn sentence representation.

Inspired by the success of contrastive learning in computer vision~\cite{zhuang2019local, tian2019contrastive, he2020momentum, chen2020simple, misra2020self}, we are interested in exploring whether it could also help language models generate a better sentence representation. The key method in contrastive learning is augmenting positive samples during the training. However, data augmentation for text is not as fruitful as for image. The image can be augmented easily by rotating, cropping, resizing, or cutouting, etc.~\cite{chen2020simple}. 
In NLP, there are minimal augmentation ways that have been researched in literature~\cite{giorgi2020declutr, fang2020cert}. 
The main reason is that every word in a sentence may play an essential role in expressing the whole meaning. Additionally, the order of the words also matters. 

Most existing pre-trained language models~\cite{devlin2019bert, liu2019roberta, lewis2019bart} are adding different kinds of noises to the text and trying to restore them at the word-level. 
Sentence-level objectives are rarely studied. BERT~\cite{devlin2019bert} combines the word-level loss, masked language modeling (MLM) with a sentence-level loss, next sentence prediction (NSP), and observes that MLM+NSP is essential for some downstream tasks. RoBERTa~\cite{liu2019roberta} drops the NSP objective during the pre-training but achieves a much better performance in a variety of downstream tasks. 
ALBERT~\cite{lan2019albert} proposes a self-supervised loss for Sentence-Order Prediction (SOP), which models the inter-sentence coherence. Their work shows that coherence prediction is a better choice than the topic prediction, the way NSP uses. 
DeCLUTR~\cite{giorgi2020declutr} is the first work to combine Contrastive Learning (CL) with MLM into pre-training. However, it requires an extremely long input document, i.e., 2048 tokens, which restricts the model to be pre-trained on limited data. Further, DeCLUTR trains from existing pre-trained models, so it remains unknown whether it could also achieve the same performance when it trains from scratch.

Drawing from the recent advances in pre-trained language models and contrastive learning, we propose a new framework, CLEAR, combining word-level MLM objective with sentence-level CL objective to pre-train a language model. MLM objective enables the model capture word-level hidden features while CL objective ensures the model with the capacity of recognizing similar meaning sentences by training an encoder to minimize the distance between the embeddings of different augmentations of the same sentence. In this paper, we present a novel design of augmentations that can be used to pre-train a language model at the sentence-level. Our main findings and contributions can be summarized as follows:

\begin{itemize}
    \item
    We proposed and tested four basic sentence augmentations: random-words-deletion, spans-deletion, synonym-substitution,  and reordering, which fills a large gap in NLP about what kind of augmentations can be used in contrastive learning.
    \item
    We showed that model pre-trained by our proposed method outperforms several strong baselines (including RoBERTa and BERT) on both GLUE~\cite{wang2018glue} and SentEval~\cite{conneau2018senteval} benchmark. For example, we showed +2.2\% absolute improvement on 8 GLUE tasks and +5.7\% absolute improvement on 7 SentEval semantic textual similarity tasks compared to RoBERTa model.
\end{itemize}

\section{Related Work} \label{related-work}
There are three lines of literatures that are closely related to our work: sentence representation, large-scale pre-trained language representation models, contrastive learning.

\subsection{Sentence Representation}
Learning the representation of sentence has been studied by many existing works. Applying various pooling strategies onto word embeddings as the representation of sentence is a common baseline~\cite{iyyer2015deep, shen2018baseline, reimers2019sentence}. 
Skip-Thoughts~\cite{kiros2015skip} trains an encoder-decoder model trying to reconstruct surrounding sentences. Quick-Thoughts~\cite{logeswaran2018efficient} trains a encoder-only model with the ability to select the correct context of the sentence out of other contrastive sentences. Later on, many pre-trained language models such as BERT~\cite{devlin2019bert} propose to use the manually-inserted token (the $\mathrm{[CLS]}$ token) as the representation of the whole sentence and become the new state-of-the-art in a variety of downstream tasks. One recent paper Sentence-BERT~\cite{reimers2019sentence} compares the average BERT embeddings with the CLS-token embedding and surprisingly finds that computing the mean of all output vectors at the last layer of BERT outperforms the CLS-token marginally.

\subsection{Large-scale Pre-trained Language Representation Models} \label{sec2.2}

The deep pre-trained language models have proven their powers in capturing implicit language features even with different model architectures, pre-training tasks, and loss functions. Two of the early works that are GPT~\cite{radford2018improving} and BERT~\cite{devlin2019bert}: GPT uses a left-to-right Transformer while BERT designs a bidirectional Transformer. Both created an incredible new state of the art in a lot of downstream tasks.

Following this observation, recently, a tremendous number of research works are published in the pre-trained language model domain. Some extend previous models to a sequence-to-sequence structure~\cite{song2019mass, lewis2019bart, liu2020multilingual}, which enforces the model's capability on language generation. The others~\cite{yang2019xlnet,liu2019roberta,clark2020electra} explore the different pre-training objectives to either improve the model's performance or accelerate the pre-training.

\subsection{Contrastive Learning}
Contrastive Learning has become a rising domain because of its significant success in various computer vision tasks and datasets. Several researchers~\cite{zhuang2019local, tian2019contrastive, misra2020self, chen2020simple} proposed to make the representations of the different augmentation of an image agree with each other and showed positive results. The main difference between these works is their various definition of image augmentation.

Researchers in the NLP domain have also started to work on finding suitable augmentation for text. CERT~\cite{fang2020cert} applies the back-translation to create augmentations of original sentences, while DeCLUTR~\cite{giorgi2020declutr} regards different spans inside one document are similar to each others. Our model differs from CERT in adopting an encoder-only structure, which decreases noise brought by the decoder. Further, unlike DeCLUTR, which only tests one augmentation and trains the model from an existing pre-trained model, we pre-train all models from scratch, which provides a straightforward comparison with the existing pre-trained models.

\tikzset{
  FARROW/.style={arrows={-{Latex[length=1.25mm, width=1.mm]}},},
  origin/.style = {regular polygon, regular polygon sides=6, fill=silver!38!platinum, minimum width=2em, align=center, inner sep=-0.5mm, outer sep=0, font=\tiny, rounded corners=1},
  aug1/.style = {regular polygon, regular polygon sides=6, fill=monte_carlo, minimum width=2em, align=center, inner sep=-0.5mm, outer sep=0, font=\tiny, rounded corners=1},
  aug2/.style = {regular polygon, regular polygon sides=6, fill=Madang, minimum width=2em, align=center, inner sep=-0.5mm, outer sep=0, font=\tiny, rounded corners=1},
}

\begin{figure*}[t]
\centering
 \begin{subfigure}[b]{0.4\textwidth}
  \centering
  \begin{tikzpicture}
    \node [origin, above of=o, align=center] (t4) at (0, 0) {Tok$_4$};
    \foreach \x/\p in {3/4,2/3,1/2}
    {
    \node [origin, left of=t\p, node distance=1cm] (t\x) {Tok$_\x$};
    }
    \node [origin, right of=t4, node distance=1cm] (t5) {Tok$_5$};
    \node [right of=t5, node distance=0.7cm] (tc) {\scriptsize$\cdots$};
    \node [origin, right of=tc, node distance=0.7cm] (tN) {Tok$_N$};
    
    \foreach \x in {3,5,N}
    {
    \node [origin, above of=t\x, node distance=1.2cm] (a\x) {Tok$_\x$};
    \node [origin, above of=a\x, node distance=1.2cm] (ao\x) {Tok$_\x$};
    }
    \foreach \x in {1,2,4}
    {
    \node [aug1, above of=t\x, node distance=1.2cm] (a\x) {Tok$_{\scaleto{\mathrm{[del]}}{2.7pt}}$};
    }
    \node [above of=tc, node distance=1.2cm] (ac) {\scriptsize$\cdots$};
    \node [above of=ac, node distance=1.2cm] (aoc) {\scriptsize$\cdots$};
    \node [aug1, above of=a4, node distance=1.2cm] (ao4) {Tok$_{\scaleto{\mathrm{[del]}}{2.7pt}}$};
    \node [aug1, above of=a1, node distance=1.2cm, xshift=0.5cm] (ao1) {Tok$_{\scaleto{\mathrm{[del]}}{2.7pt}}$};
    
    \foreach \x in {3,5,N}
    {
    \draw[FARROW]  (t\x) ->  (a\x)  {} ;
    \draw[FARROW]  (a\x) ->  (ao\x)  {} ;
    }
    \draw[FARROW, flamingo]  (t4) ->  (a4)  {} ;
    \draw[FARROW, flamingo]  (a4) ->  (ao4)  {} ;
    \draw[FARROW, flamingo]  (t1) ->  (a1)  {} ;
    \draw[FARROW, flamingo]  (t2) ->  (a2)  {} ;
    \draw[FARROW, flamingo]  (a2.north) ->  (ao1)  {} ;
    \draw[FARROW, flamingo] (a1.north) ->  (ao1)  {} ;

    \draw [thick,draw=black!42, decorate,decoration={brace,amplitude=8pt}] ([xshift=-1mm, yshift=0.5mm] ao1.north west) -- ([xshift=1mm, yshift=0.5mm] aoN.north east) node[midway,yshift=0.5cm] (cv_a) {\small sentence after word deletion};
    \draw [thick,draw=black!42, decorate,decoration={brace,amplitude=8pt, mirror}] ([xshift=-1mm, yshift=-0.5mm] t1.south west) -- ([xshift=1mm, yshift=-0.5mm] tN.south east) node[midway,yshift=-0.5cm] (cv_b) {\small original sentence};
    
  \end{tikzpicture}

   \caption{\textbf{Word Deletion}: $\mathrm{Tok}_{1}$, $\mathrm{Tok}_{2}$, and  $\mathrm{Tok}_{4}$ are deleted, the sentence after augmentation will be: $[\mathrm{Tok}_{\text{[del]}}, \mathrm{Tok}_{3}, \mathrm{Tok}_{\text{[del]}}, \mathrm{Tok}_{5}, \dots, \mathrm{Tok}_{N}]$.}
    \label{fig:del-eli}
 \end{subfigure}
 \hfil
  \begin{subfigure}[b]{0.4\textwidth}
  \centering
  \begin{tikzpicture}
    \node [origin, above of=o, align=center] (t4) at (0, 0) {Tok$_4$};
    \foreach \x/\p in {3/4,2/3,1/2}
    {
    \node [origin, left of=t\p, node distance=1cm] (t\x) {Tok$_\x$};
    }
    \node [origin, right of=t4, node distance=1cm] (t5) {Tok$_5$};
    \node [right of=t5, node distance=0.7cm] (tc) {\scriptsize$\cdots$};
    \node [origin, right of=tc, node distance=0.7cm] (tN) {Tok$_N$};
    
    \foreach \x in {1,2,3,4}
    {
    \node [aug1, above of=t\x, node distance=1.2cm] (a\x) {Tok$_{\scaleto{\mathrm{[del]}}{2.7pt}}$};
    }
    \node [above of=tc, node distance=1.2cm] (ac) {\scriptsize$\cdots$};
    \node [above of=ac, node distance=1.2cm] (aoc) {\scriptsize$\cdots$};
    \foreach \x in {5,N}
    {
    \node [origin, above of=t\x, node distance=1.2cm] (a\x) {Tok$_\x$};
    \node [origin, above of=a\x, node distance=1.2cm] (ao\x) {Tok$_\x$};
    }
    \foreach \x in {5,N}
    {
    \draw[FARROW]  (t\x) ->  (a\x)  {} ;
    }
    
    \node [aug1, above of=a2, node distance=1.2cm, xshift=0.5cm] (ao1) {Tok$_{\scaleto{\mathrm{[del]}}{2.7pt}}$};

    \draw[FARROW]  (a5) ->  (ao5)  {} ;
    \draw[FARROW]  (aN) ->  (aoN)  {} ;
    
    \draw [decorate, flamingo, decoration={brace,amplitude=5pt}] ([xshift=-1mm, yshift=1.2mm] a1.north west) -- ([xshift=1mm, yshift=1.2mm] a4.north east) node[midway,yshift=0.5cm] (a_a) {};
    \draw[FARROW, flamingo]  ([yshift=-4.5mm] a_a.north) ->  (ao1.south)  {} ;
    
    \node [thick, dotted, draw=flamingo, fit={(t1) (t4)}, inner sep=1, rounded corners=2] (span1) {};
    \node [thick, dotted, draw=flamingo, fit={(a1) (a4)}, inner sep=1, rounded corners=2] (span2) {};
     \draw[FARROW, flamingo]  (span1) ->  (span2)  {} ;

    \draw [thick,draw=black!42, decorate,decoration={brace,amplitude=8pt}] ([xshift=-1mm, yshift=0.5mm] ao1.north west) -- ([xshift=1mm, yshift=0.5mm] aoN.north east) node[midway,yshift=0.5cm] (cv_a) {\small sentence after span deletion};
    \draw [thick,draw=black!42, decorate,decoration={brace,amplitude=8pt, mirror}] ([xshift=-1mm, yshift=-1mm] t1.south west) -- ([xshift=1mm, yshift=-1mm] tN.south east) node[midway,yshift=-0.5cm] (cv_b) {\small original sentence};
    
  \end{tikzpicture}

   \caption{\textbf{Span Deletion}: The span $[\mathrm{Tok}_1, \mathrm{Tok}_2, \mathrm{Tok}_3$, $\mathrm{Tok}_4]$ is deleted, the sentence after augmentation will be: $[\mathrm{Tok}_{\text{[del]}}, \mathrm{Tok}_{5}, \dots, \mathrm{Tok}_{N}]$.}
    \label{fig:del-span}
 \end{subfigure}
\par\bigskip
\begin{subfigure}[t]{0.4\textwidth}
  \centering
  \begin{tikzpicture}
    \node [origin, above of=o, align=center] (t4) at (0, 0) {Tok$_4$};
    \foreach \x/\p in {3/4,2/3,1/2}
    {
    \node [origin, left of=t\p, node distance=1cm] (t\x) {Tok$_\x$};
    }
    \node [origin, right of=t4, node distance=1cm] (t5) {Tok$_5$};
    \node [right of=t5, node distance=0.7cm] (tc) {\scriptsize$\cdots$};
    \node [origin, right of=tc, node distance=0.7cm] (tN) {Tok$_N$};
    \node [above of=tc, node distance=1.5cm] (ac) {\scriptsize$\cdots$};

    \foreach \x/\p in {2/3, 5/5,N/N}
    {
    \node [origin, above of=t\x, node distance=1.5cm] (a\p) {Tok$_\p$};
    }
    \foreach \x/\p in {1/4,3/1,4/2}
    {
    \node [aug1, above of=t\x, node distance=1.5cm] (a\p) {Tok$_\p$};
    }
    \node [thick, dotted, draw=flamingo, fit={(t1) (t2)}, inner sep=1] (span1) {};
    \node [thick, dotted, draw=flamingo, fit={(t4)}, inner sep=1] (span4) {};
    \node [thick, dotted, draw=flamingo, fit={(a1) (a2)}, inner sep=1] (span2) {};
    \node [thick, dotted, draw=flamingo, fit={(a4)}, inner sep=1] (span3) {};
    
    \draw[FARROW, flamingo]  (span1.north) ->  (span2.south)  {} ;
    \draw[FARROW, flamingo]  (span4.north) ->  (span3.south)  {} ;
    \foreach \x in {3,5,N}
    {
    \draw[FARROW]  (t\x.north) ->  (a\x.south)  {} ;
    }
    \draw [thick,draw=black!42, decorate,decoration={brace,amplitude=8pt}] ([xshift=-1mm, yshift=0.8mm] a4.north west) -- ([xshift=1mm, yshift=0.8mm] aN.north east) node[midway,yshift=0.5cm] (cv_a) {\small sentence after reordering};
    \draw [thick,draw=black!42, decorate,decoration={brace,amplitude=8pt, mirror}] ([xshift=-1mm, yshift=-0.8mm] t1.south west) -- ([xshift=1mm, yshift=-0.8mm] tN.south east) node[midway,yshift=-0.5cm] (cv_b) {\small original sentence};
    
  \end{tikzpicture}

   \caption{\textbf{Reordering}: Two spans $[\mathrm{Tok}_1,\, \mathrm{Tok}_2]$ and $[\mathrm{Tok}_4]$ are reordered, the sentence after augmentation will be: $[\mathrm{Tok}_4,\, \mathrm{Tok}_3,\, \mathrm{Tok}_1,\, \mathrm{Tok}_2,$ $\mathrm{Tok}_5, \dots, \mathrm{Tok}_{N}]$.}
    \label{fig:reorder}
 \end{subfigure}
 \hfil
 \begin{subfigure}[t]{0.4\textwidth}
  \centering
  \begin{tikzpicture}
    \node [origin, above of=o, align=center] (t4) at (0, 0) {Tok$_4$};
    \foreach \x/\p in {3/4,2/3,1/2}
    {
    \node [origin, left of=t\p, node distance=1cm] (t\x) {Tok$_\x$};
    }
    \node [origin, right of=t4, node distance=1cm] (t5) {Tok$_5$};
    \node [right of=t5, node distance=0.7cm] (tc) {\scriptsize$\cdots$};
    \node [origin, right of=tc, node distance=0.7cm] (tN) {Tok$_N$};

    \foreach \x in {1,4,5}
    {
    \node [origin, above of=t\x, node distance=1.5cm] (a\x) {Tok$_\x$};
    }
    \node [above of=tc, node distance=1.5cm] (ac) {\scriptsize$\cdots$};
    \foreach \x in {2,3,N}
    {
    \node [aug1, above of=t\x, node distance=1.5cm] (a\x) {Tok$_\x'$};
    }
     \foreach \x in {1,4,5}
    {
    \draw[FARROW]  (t\x) ->  (a\x)  {} ;
    }
    \foreach \x in {2,3,N}
    {
    \draw[FARROW, flamingo]  (t\x) ->  (a\x)  {} ;
    }
    \draw [thick,draw=black!42, decorate,decoration={brace,amplitude=8pt}] ([xshift=-1mm, yshift=0.8mm] a1.north west) -- ([xshift=1mm, yshift=0.8mm] aN.north east) node[midway,yshift=0.5cm] (cv_a) {\small sentence after similar word subsitution};
    \draw [thick,draw=black!42, decorate,decoration={brace,amplitude=8pt, mirror}] ([xshift=-1mm, yshift=-0.8mm] t1.south west) -- ([xshift=1mm, yshift=-0.8mm] tN.south east) node[midway,yshift=-0.5cm] (cv_b) {\small original sentence};
    
  \end{tikzpicture}

   \caption{\textbf{Synonym Substitution}: $\mathrm{Tok}_2$, $\mathrm{Tok}_3$, and $\mathrm{Tok}_N$ are substituted by their synonyms $\mathrm{Tok}_2'$, $\mathrm{Tok}_3'$, and $\mathrm{Tok}_N'$, respectively. The sentence after augmentation will be: $[\mathrm{Tok}_1,\, \mathrm{Tok}_2',\, \mathrm{Tok}_3', \mathrm{Tok}_4,$ $\mathrm{Tok}_5, \dots, \mathrm{Tok}_N']$.}
    \label{fig:subs}
 \end{subfigure}
 
\caption{Four sentence augmentation methods in proposed contrastive learning framework CLEAR.}
\label{fig:augmentations}
\end{figure*}
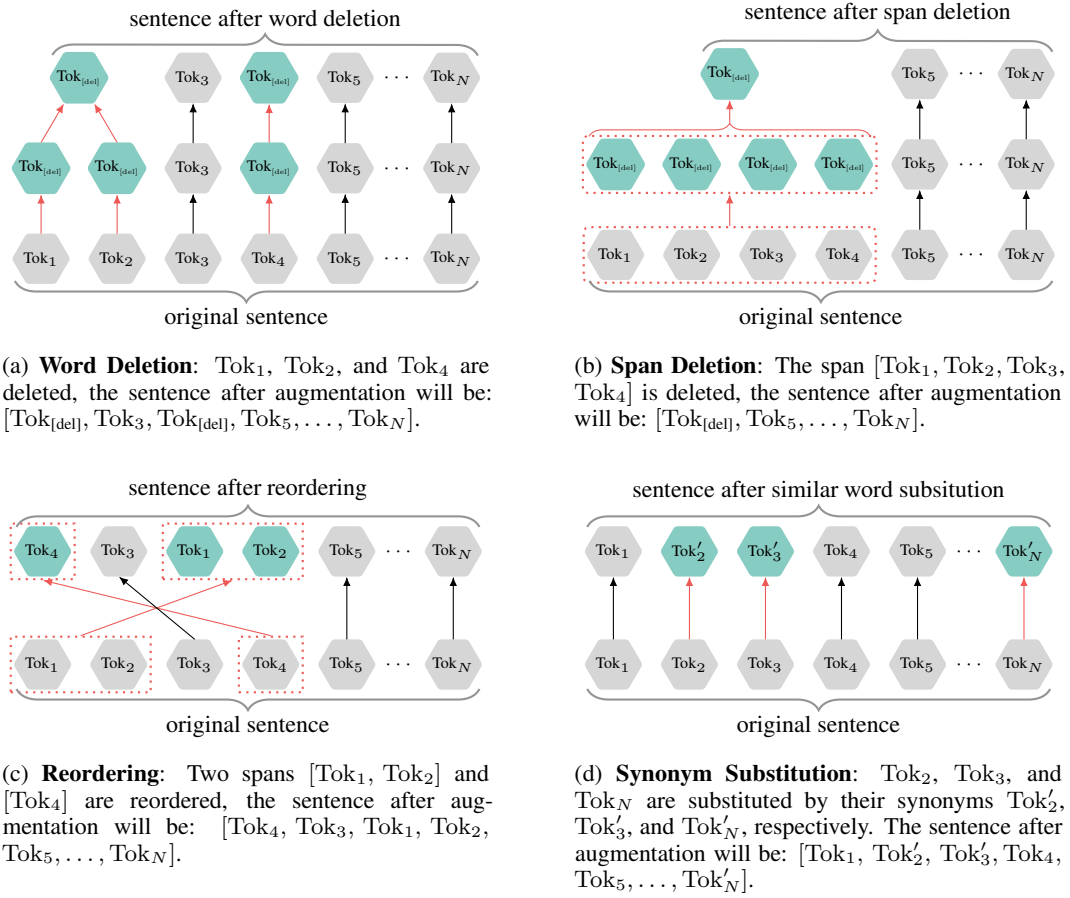

\section{Method} \label{method}
This section proposes a novel framework and several sentence augmentation methods for contrastive learning in NLP. 

\subsection{The Contrastive Learning Framework} 
\label{3_1}
Borrow from SimCLR~\cite{chen2020simple}, we propose a new contrastive learning framework to learn the sentence representation, named as CLEAR. There are four main components in CLEAR, as outlined in Figure~\ref{fig:cl_model}.
\begin{itemize}
    \item An augmentation component $\mathrm{AUG}(\cdot)$ which apply the random augmentation to the original sentence. For each original sentence $s$, we generate two random augmentations $\widetilde{s}_1 = \mathrm{AUG}(s, \mathit{seed}_1)$ and $\widetilde{s}_2 = \mathrm{AUG}(s, \mathit{seed}_2)$, where $\mathit{seed}_1$ and $\mathit{seed}_2$ are two random seeds. 
    Note that, to test each augmentation's effect solely, we adopt the same augmentation to generate $\widetilde{s}_1$ and $\widetilde{s}_2$. Testing the mixing augmentation models requests more computational resources, which we plan to leave for future work.
    We will detail the proposed augmentation set $\mathcal{A}$ at Section \ref{3_3}. 
    \item A transformer-based encoder $f(\cdot)$ that learns the representation of the input augmented sentences $H_1 = f(\widetilde{s}_1)$ and $H_2 = f(\widetilde{s}_2)$. Any encoder that learns the sentence representation can be used here to replace our encoder. We choose the current start-of-the-art (i.e., transformer~\cite{vaswani2017attention}) to learn sentence representation and use the representation of a manually-inserted token as the vector of the sentence (i.e., $\mathrm{[CLS]}$, as used in BERT and RoBERTa). 
    \item A nonlinear neural network projection head $g(\cdot)$ that maps the encoded augmentations $H_1$ and $H_2$ to the vector $z_1=g(H_1)$, $z_2=g(H_2)$ in a new space. According to observations in SimCLR~\cite{chen2020simple}, adding a nonlinear projection head can significantly improve representation quality of images.
    \item A contrastive learning loss function defined for a contrastive prediction task, i.e., trying to predict positive augmentation pair ($\widetilde{s}_1$, $\widetilde{s}_2$) in the set $\{\tilde{s}\}$. We construct the set $\{\tilde{s}\}$ by randomly augmenting twice for all the sentences in a minibatch (assuming a minibatch is a set $\{s\}$ size $N$), getting a set $\{\tilde{s}\}$ with size $2N$. The two variants from the same original sentence form the positive pair, while all other instances from the same minibatch are regarded as negative samples for them. The contrastive learning loss has been tremendously used in previous work~\cite{wu2018unsupervised, chen2020simple, giorgi2020declutr, fang2020cert}. The loss function for a positive pair is defined as:
    \begin{equation}
        l(i, j) {=} {-}\log \frac{\exp\left(\mathrm{sim}(z_i, z_j) / {\tau}\right)}{\sum_{k=1}^{2N} \mathbbm{1}_{[k\neq i]}\exp\left(\mathrm{sim}(z_i, z_k) / {\tau}\right) }
    \end{equation}
    where $\mathbbm{1}_{[k\neq i]}$ is the indicator function to judge whether $k \neq i$, $\tau$ is a temperature parameter, $\mathrm{sim}(u, v) = u^{\top}v/(\Vert u\Vert_2 \Vert v\Vert_2)$ denotes the cosine similarity of two vector $u$ and $v$. The overall contrastive learning loss is defined as the sum of all positive pairs' loss in a mini-batch:
    \begin{equation}
        \mathcal{L}_{\mathrm{CL}} = \sum_{i=1}^{2N} \sum_{j=1}^{2N} m(i, j) l(i,j)
    \end{equation}
    where $m(i,j)$ is a function returns $1$ when $i$ and $j$ is a positive pair, returns $0$ otherwise.  
\end{itemize}

\subsection{The Combined Loss for Pre-training} \label{3_2}
Similar to~\cite{giorgi2020declutr}, for the purpose of grabbing both token-level and sentence-level features, we use a combined loss of MLM objective and CL objective to get the overall loss:
\begin{equation}
    \mathcal{L}_{\mathrm{total}} = \mathcal{L}_{\mathrm{MLM}} + \mathcal{L}_{\mathrm{CL}}
\end{equation}
where $\mathcal{L}_{\mathrm{MLM}}$ is calculated through predicting the random-masked tokens in set $\{s\}$ as described in BERT and RoBERTa~\cite{devlin2019bert, liu2019roberta}. Our pre-training target is to minimize the $\mathcal{L}_{\mathrm{total}}$.

\subsection{Design Rationale for Sentence Augmentations} 
\label{3_3}
The data augmentation is crucial for learning the representation of image~\cite{tian2019contrastive, jain2020contrastive}. However, in language modeling, it remains unknown whether data (sentence) augmentation would benefit the representation learning and what kind of data augmentation could apply to the text. 
To answer these questions, we explore and test four basic augmentations (shown in Figure \ref{fig:augmentations}) and their combinations in our experiment. We do believe there exist more potential augmentations, which we plan to leave for future exploration. 

One type of augmentation we consider is \textbf{deletion}, which bases on the hypothesis that some deletion in a sentence wouldn't affect too much of the original semantic meaning. In some case, it may happen that deleting some words leads the sentence to a different meaning (e.g., the word \textit{not}). However, we believe including proper noise can benefit the model to be more robust. We consider two different deletions, i.e., \textbf{word deletion} and \textbf{span deletion}. 
\begin{itemize}
    \item Word deletion (shown in Figure \ref{fig:del-eli}) randomly selects tokens in the sentence and replace them by a special token $\mathrm{[DEL]}$, which is similar to the token $\mathrm{[MASK]}$ in BERT~\cite{devlin2019bert}.
    \item Span deletion (shown in Figure \ref{fig:del-span}) picks and replaces the deletion objective on the span-level. Generally, span-deletion is a special case of word-deletion, which puts more focus on deleting consecutive words.
\end{itemize}
To avoid the model easily distinguishing the two augmentations from the remaining words at the same location, we eliminate the consecutive token $\mathrm{[DEL]}$ into one token. 

\textbf{Reordering} (shown in  Figure \ref{fig:reorder}) is another widely-studied augmentation that can keep the original sentence's features. BART~\cite{lewis2019bart} has explored restoring the original sentence from the random reordered sentence. We randomly sample several pairs of span and switch them pairwise to construct the reordering augmentation in our implementation. 

\textbf{Substitution} (shown in Figure \ref{fig:subs}) has been proven efficient in improving model's robustness~\cite{jia2019certified}. Following their work, we sample some words and replace them with synonyms to construct one augmentation. The synonym list comes from a vocabulary they used. In our pre-training corpus, there are roughly $40\%$ tokens with at least one similar-meaning token in the list. 

\section{Experiment} \label{experiment}
This section presents empirical experiments that compare the proposed methods with various baselines and alternative approaches.
\begin{table*}[t]
\caption{Performance of competing methods evaluated on GLUE dev set. Following GLUE's setting~\cite{wang2018glue}, unweighted average accuracy on the matched and mismatched dev sets is reported for MNLI. The unweighted average of accuracy and F1 is reported for MRPC and QQP. The unweighted average of Pearson and Spearman correlation is reported for STS-B. The Matthews correlation is reported for CoLA. For all other tasks we report accuracy.} 
\label{tab:large-glue}
\centering
\setlength{\tabcolsep}{8pt}
\renewcommand{\arraystretch}{1.2}
\begin{adjustbox}{max width=\textwidth}
\begin{tabular}{l c c c c c c c c c}  %
\toprule
Method & MNLI & QNLI & QQP & RTE & SST-2 & MRPC & CoLA & STS & Avg \\
\midrule
Baselines & & & & & & & & & \\
BERT-base~\cite{devlin2019bert} & 84.0 & 89.0 & 89.1 & 61.0 & 93.0 & 86.3 & 57.3 & 89.5 & 81.2\\
RoBERTa-base~\cite{liu2019roberta} & 87.2 & 93.2 & 88.2 & 71.8 & 94.4 & 87.8 & 56.1 & 89.4 & 83.5 \\
\midrule
MLM+1-CL-objective & & & & & & & & & \\
\textbf{MLM+ del-word} & 86.8 & 93.0 & 90.2 & 79.4 & 94.2 & 89.7  & 62.1 & \textbf{90.5} & \textbf{85.7}\\
\textbf{MLM+ del-span} & \textbf{87.3} & 92.8 & 90.1 & \textbf{79.8} & 94.4 & 89.9 & 59.8 & 90.3 & 85.6\\
\midrule
MLM+2-CL-objective & & & & & & & & & \\
\textbf{MLM+ subs+ del-word} & \textbf{87.3} & 93.1 & 90.0 & 73.3 & 93.7 & 90.2 & 62.1 & 90.1 & 85.0\\
\textbf{MLM+ subs+ del-span} & 87.0 & \textbf{93.4} & \textbf{90.3} & 74.4  & 94.3 & 90.5 & 63.3 & \textbf{90.5} & 85.5\\
\textbf{MLM+ del-word+ reorder} & 87.0 & 92.7 & 89.5 & 76.5 & \textbf{94.5} & \textbf{90.6} & 59.1 & 90.4 & 85.0\\
\textbf{MLM+ del-span+ reorder} & 86.7 & 92.9 & 90.0 & 78.3 & \textbf{94.5} & 89.2 & \textbf{64.3} & 89.8 & \textbf{85.7}\\

\bottomrule
\end{tabular}

\end{adjustbox}
\end{table*}

\subsection{Setup} \label{exp-setup}
\noindent \textbf{Model configuration:} We use the Transformer (12 layers, 12 heads and 768 hidden size) as our primary encoder~\cite{vaswani2017attention}. Models are pre-trained for 500K updates, with mini-batches containing 8,192 sequences of maximum length 512 tokens. For the first 24,000 steps, the learning rate is warmed up to a peak value of $6\mathrm{e}{-4}$, then linearly decayed for the rest. All models are optimized by Adam~\cite{kingma2014adam} with $\beta_1 = 0.9, \beta_2 = 0.98, \epsilon = 1\mathrm{e}{-6}$, and $L_2$ weight decay of $0.01$. We use $0.1$ for dropout on all layers and in attention. All of the models are pre-trained on 256 NVIDIA Tesla V100 32GB GPUs.

\noindent \textbf{Pre-training data:} We pre-train all the models on a combination of BookCorpus~\cite{zhu2015aligning} and English Wikipedia datasets, the data BERT used for pre-training. For more statistics of the dataset and processing details, one can refer to BERT~\cite{devlin2019bert}.

\noindent \textbf{Hyperparameters for MLM:} For calculating MLM loss, we randomly mask $15\%$ tokens of the input text \textbf{s} and use the surrounding tokens to predict them. To fill the gap between fine-tuning and pre-training, we also adopt the $10\%$-random-replacement and $10\%$-keep-unchanged setting in BERT for the masked tokens.

\noindent \textbf{Hyperparameters for CL:} To compute CL loss, we set up different hyperparameters: 
\begin{itemize}
  \item For \textbf{Word Deletion (del-word)}, we delete $70\%$ tokens.
  \item For \textbf{Span Deletion (del-span)}, we delete 5 spans (each with $5\%$ length of the input text).
  \item For \textbf{Reordering (reorder)}, we randomly pick 5 pairs of spans (each with roughly $5\%$ length as well) and switch spans pairwise.
  \item For \textbf{Substitution (subs)}, we randomly select $30\%$ tokens and replace each token with one of their similar-meaning tokens.
\end{itemize}   
Some of the above hyperparameters are slightly-tuned on the WiKiText-103 dataset~\cite{merity2016pointer} (trained for 100 epochs, evaluated on the GLUE dev benchmark). For example, we find $70\%$ deletion model perform best out of $\{30\%, 40\%, 50\%, 60\%, 70\%, 80\%, 90\%\}$ deletion models. For models using mixed augmentations, like MLM+2-CL-objective in Table \ref{tab:large-glue}, they use the same optimized hyperparameters as in the single model. For instance, our notation MLM+subs+del-span represents a model combining the MLM loss with CL loss: for MLM, it masks $15\%$ tokens; for CL, it substitutes $30\%$ tokens first and then deletes 5 spans to generate augmented sentences. 

Note that the hyperparameters we used might not be the most optimized ones. Yet, it is unknown whether optimized hyperparameters on a 1-CL-objective model perform consistently on a 2-CL-objective model. Additionally, it is also unclear whether the optimized hyperparameters for WiKiText-103 are still the optimized ones on BookCorpus and English Wikipedia datasets. However, it is hard to tune every possible hyperparameter due to the extensive computation resource requirement for pre-training. We will leave these questions to explore in the future. 

\begin{table*}[t]
\caption{Performance of competing methods evaluated on SentEval. All results are pre-trained on BookCorpus and English Wikipedia datasets for 500k steps.}
\label{tab:senteval}
\centering
\setlength{\tabcolsep}{8pt}
\renewcommand{\arraystretch}{1.2}
\begin{adjustbox}{max width=\textwidth}

\begin{tabular}{l c c c c c c c c}  %
\toprule
Method & SICK-R & STS-B &  STS12 & STS13 & STS14 & STS15 & STS16 & Avg\\
\midrule
Baselines & & & & & & & &\\
RoBERTa-base-mean & 74.1 & 65.6 & 47.2 & 38.3 & 46.7 & 55.0 & 49.5 & 53.8\\
RoBERTa-base-[CLS] & 75.9 & 71.9 & 47.4 & 37.5 & 47.9 & 55.1 & 57.6 & 56.1\\
\midrule
MLM+1-CL-objective & & & & & & & &\\
\textbf{MLM+ del-word-mean} & 75.9 & 69.0 & \textbf{50.6} & 40.0 & 50.2 & 58.9 & 52.4 & 56.7\\
\textbf{MLM+ del-span-mean} & 71.0 & 62.6 & 49.3 & 41.7 & 48.9 & 58.1 & 52.3 & 54.8\\
\textbf{MLM+ del-word-[CLS]} & \textbf{77.1} & 71.6 & \textbf{50.6} & 44.5 & 48.3 & 58.4 & 56.1 & 58.1\\
\textbf{MLM+ del-span-[CLS]} & 62.7 & 57.4 & 34.4 &  20.4 & 24.3 & 32.0  & 31.5 & 37.5\\
\midrule
MLM+2-CL-objective & & & & & & & &\\
\textbf{MLM+ del-word+ reorder-mean} & 75.8 & 66.2 & 51.1 & 45.7 & 51.8 & 61.3  & 57.0 & 58.4\\
\textbf{MLM+ del-span+ reorder-mean} & 75.4 & 67.8 & 48.3 & 50.3 & 54.9 & 60.4 & 56.8 & 59.1\\
\textbf{MLM+ subs+ del-word-mean} & 73.6 & 63.4 & 44.6 & 39.8 & 50.1 & 55.5 & 49.6 & 53.8\\
\textbf{MLM+ subs+ del-span-mean} & 75.5 & 67.0 & 48.3 & 45.0 & 54.6 & 60.9 & 58.5 & 58.5\\
\textbf{MLM+ del-word+ reorder-[CLS]} & 71.9 & 63.8 & 41.9 & 30.9 & 37.4 & 48.9 & 52.1 & 49.6\\
\textbf{MLM+ del-span+ reorder-[CLS]} & 75.0 & 68.7 & 49.4 & \textbf{54.3} & \textbf{57.6} & \textbf{64.0} & 61.4 & 61.5\\
\textbf{MLM+ subs+ del-word-[CLS]} & 73.6 & 62.9 & 44.5 & 35.8 & 47.6 & 55.8 & 59.6  & 54.3\\
\textbf{MLM+ subs+ del-span-[CLS]} & 75.6 & \textbf{72.5} & 49.0 & 48.9 & 57.4 & 63.6 & \textbf{65.6} & \textbf{61.8} \\
\bottomrule
\end{tabular}

\end{adjustbox}
\end{table*}

\subsection{GLUE Results} \label{exp-glue-wiki103}
We mainly evaluate all the models by the General Language Understanding Evaluation (GLUE) benchmark development set~\cite{wang2018glue}. GLUE is a benchmark containing several different types of NLP tasks: natural language inference task (MNLI, QNLI, and RTE), similarity task (QQP, MRPC, STS), sentiment analysis task (SST), and linguistic acceptability task(CoLA). It provides a comprehensive evaluation for pre-trained language models.

To fit the different downstream tasks' requirements, we follow the RoBERTa's hyperparamters to finetune our model for various tasks. Specifically, we add an extra fully connected layer and then finetune the whole model on different training sets. 

The primary baselines we include are BERT-base and RoBERTa-base.  The results for BERT-base are from huggingface's reimplementation\footnote{https://huggingface.co/transformers/v1.1.0/examples.html}. A more fair comparison comes from RoBERTa-base since we use the same hyperparameters RoBERTa-base used for MLM loss. Note that our models are all combining two-loss, it is still unfair to compare a MLM-only model with a MLM+CL model. To answer this question, we set two other baselines in Section ~\ref{5_1} to make a more strict comparison: one combines two MLM losses, the other adopts a double batch size. 

As we can see in Table \ref{tab:large-glue}, our proposed several models outperform the baselines on GLUE. Note that different tasks adopt different evaluation matrices, our two best models MLM+del-word and MLM+del-span+reorder both improve the best baseline RoBERTa-base by 2.2\% on average score. Besides, a more important observation is that all best performance for each task comes from our proposed model. On CoLA and RTE, our best model exceeds the baseline by 7.0\% and 8.0\% correspondingly. Further, we also find that different downstream tasks benefit from different augmentations. We will make a more specific analysis in Section \ref{5_2}.

One notable thing is that we don't show the result of MLM+subs, MLM+reorder, and MLM+subs+reorder in Table \ref{tab:large-glue}. We observe that the pre-training for these three models either converges quickly or suffers from a gradient explosion problem, which indicates that these three augmentations are too easy to distinguish.

\begin{table*}[t]
\caption{Ablation study for several methods evaluated on GLUE dev set. All results are pre-trained on wiki-103 data for 500 epochs.}
\label{tab:dis}
\centering
\setlength{\tabcolsep}{8pt}
\renewcommand{\arraystretch}{1.2}
\begin{adjustbox}{max width=\textwidth}
\begin{tabular}{l c c c c c c c c c}  %
\toprule
Method & MNLI-m & QNLI & QQP & RTE & SST-2 & MRPC & CoLA & STS & Avg \\
\midrule
RoBERTa-base & 80.4 & 87.5 & 87.4 & 61.4 & 91.4 & \textbf{82.4} & 38.9 & 81.9 & 76.4 \\
\midrule
MLM-variant & & & & & & & & & \\
Double-batch RoBERTa-base & 80.3 & 88.0 & 87.1 & 59.9 & 91.9 & 82.1 & 43.0 & 82.0 & 76.8 \\
Double MLM RoBERTA-base & 80.5 & 87.6 & 87.3 & 57.4 & 90.4 & 77.7 & 42.2 & 83.0 & 75.8\\
\midrule
MLM+CL-objective & & & & & & & & & \\
\textbf{MLM+ del-span} & 80.6 & \textbf{88.8} & 87.3 & \textbf{62.1} & \textbf{92.1} & 77.8 & 44.1 & 81.4 & 76.8\\
\textbf{MLM+ del-span + reorder} & \textbf{81.1} & 88.7 & \textbf{87.5} & 58.1 & 90.0 & 80.4 & 43.3 & \textbf{87.4} & 77.1\\
\textbf{MLM+ subs + del-word + reorder} & 80.5 & 87.7 & 87.3 & 59.6 & 90.4 & 80.2 & \textbf{45.1} & 87.1 & \textbf{77.2}\\
\bottomrule
\end{tabular}

\end{adjustbox}
\end{table*}

\subsection{SentEval Results for Semantic Textual Similarity Tasks} \label{exp-senteval}
SentEval is a popular benchmark for evaluating general-purpose sentence representations~\cite{conneau2018senteval}. The specialty for this benchmark is that it doesn't do the fine-tuning like in GLUE. We evaluate the performance of our proposed methods for common  Semantic Textual Similarity (STS) tasks on SentEval. 
Note that some previous models (e.g., Sentence-BERT~\cite{reimers2019sentence}) on the SentEval leaderboard trains on the specific datasets such as Stanford NLI~\cite{bowman2015large} and MultiNLI~\cite{williams2017broad}, which makes it hard for a direct comparison. To make it easier, we compare one of our proposed models with RoBERTa-base directly on SentEval. According to Sentence-BERT, using the mean of all output vectors in the last layer is more effective than using the CLS-token output. We test both pooling strategies for each model. 

From Table \ref{tab:senteval}, we observe that mean-pooling strategy does not show much advantages. In many of the cases, CLS-pooling is better than the mean-pooling for our proposed models. The underlying reason is that the contrastive learning directly updates the representation of [CLS] token. 
Besides that, we find adding the CL loss makes the model especially good at the Semantic Textual Similarity (STS) task, beating the best baseline by a large margin (+5.7\%). We think it is because the pre-training of contrastive learning is to find the similar sentence pairs, which aligns with STS task. This could explain why our proposed models show such large improvements on STS. 
\section{Discussion} \label{discussion}
This section discusses an ablation study to compare the CL loss and MLM loss and shows some observations about what different augmentation learns. 

\subsection{Ablation Study} \label{5_1}
Our proposed CL-based models outperforms MLM-based models, one remaining question is, where does our proposed model benefit from? Does it come from the CL loss, or is it from the larger batch size (since to calculate CL loss, one needs to store extra information per batch)? To answer this question, we set up two extra baselines: Double MLM RoBERTa-base adopts the MLM+MLM loss, each MLM is performed on different mask for the same original sentence; the other Double-batch RoBERTa-base uses single MLM loss with a double-size batch. 

Due to the limitation of computational resource, we conduct the ablation study on a smaller pre-training corpus, i.e., WiKiText-103 dataset~\cite{merity2016pointer}. All the models listed in Table~\ref{tab:dis} are pre-trained for 500 epochs on 64 NVIDIA Tesla V100 32GB GPUs. Three of our proposed models are reported in the table. 
The general performance for the variants doesn't show much difference compared with the original RoBERTa-base, with a +0.4\% increase on the average score on Double-batch RoBERTa-base, which confirms the idea that a larger batch benefits the representation training as proposed by previous work~\cite{liu2019roberta}. Yet, the best-performed baseline is still not as good as our best-proposed model. It tells us the proposed model does not solely benefit from a larger batch; CL loss also helps.

\subsection{Different Augmentation Learns Different Features} \label{5_2}
In Table \ref{tab:large-glue}, we find an interesting phenomenon: different proposed models are good at specific tasks.

One example is MLM+subs+del-span helps the model be good at dealing with similarity and paraphrase tasks. On QQP and STS, it achieves the highest score; on MRPC, it ranks second. We infer the outperformance of MLM+subs+del-span in this kind of task is because synonym substitution helps translate the original sentence to similar meaning sentences while deleting different spans makes more variety of similar sentences visible. Combining them enhances the model's capacity to deal with many unseen sentence pairs. 

We also notice that MLM+del-span achieves good performance on inference tasks (MNLI, QNLI, RTE). The underlying reason is, with a span deletion, the model has already been pre-trained well to infer the other similar sentences.
The ability to identify similar sentence pairs helps to recognize the contradiction. 
Therefore, the gap between the pre-trained task and this downstream task narrows.

Overall, we observe that different augmentation learns different features. Some specific augmentations are especially good at some certain downstream tasks. Designing a task-specific augmentation or exploring meta-learning to adaptively select different CL objectives is a promising future direction. 

\section{Conclusion} \label{conclusion}
In this work, we presented an instantiation for contrastive sentence representation learning. By carefully designing and testing different data augmentations and combinations, we prove the proposed methods' effectiveness on GLUE and SentEval benchmark under the diverse pre-training corpus. 

The experiment results indicate that the pre-trained model would be more robust when leveraging adequate sentence-level supervision. More importantly, we reveal that different augmentation learns different features for the model. Finally, we demonstrate that the performance improvement comes from both the larger batch size and the contrastive loss.

\bibliography{reference}
\bibliographystyle{acl_natbib}

\end{document}